\documentclass{article}
\usepackage{amssymb}

\usepackage{amsmath}
\usepackage{jmlr2e}
\usepackage{graphics}
\usepackage{graphicx}

\newtheorem{lemma}{Lemma}[section]
\newtheorem{definition}{Definition}[section]

\begin{document}

\title{Concentration inequalities of the cross-validation estimator for
Empirical Risk Minimiser}
\author{\name  Matthieu Cornec \\
%EndAName
\addr  CREST-INSEE  \\
\addr  15 Boulevard Gabriel Peri \\
\addr  Timbre  G120\\
\addr 92240 MALAKOFF - FRANCE}

\editor{XXX}

\maketitle

\begin{abstract}
\bigskip

In this article, we derive concentration inequalities for the
cross-validation estimate of the generalization error for
empirical risk minimizers. In the general setting, we prove
sanity-check bounds in the spirit of \cite{KR99}
\textquotedblleft\textit{bounds showing that the worst-case error
of this estimate is not much worse that of training error
estimate} \textquotedblright . General loss functions and class of
predictors with finite VC-dimension are considered. We closely follow the formalism introduced by \cite{DUD03} to cover a
large variety of
cross-validation procedures including leave-one-out cross-validation, $k$%
-fold cross-validation, hold-out cross-validation (or split sample), and the
leave-$\upsilon$-out cross-validation.

\bigskip

\noindent  In particular, we focus on proving the consistency of
the various cross-validation procedures. We point out the
interest of each cross-validation procedure in terms of rate of
convergence. An estimation curve with transition phases depending
on the cross-validation procedure and not only on the percentage
of observations in the test sample gives a simple rule on how to
choose the cross-validation. An interesting consequence is that
the size of the test sample is not required to grow to infinity
for the consistency of the cross-validation procedure.

\bigskip

\end{abstract}

\addcontentsline{toc}{section}{Introduction}
\markboth{\uppercase
{Introduction}} {\uppercase{Introduction}}

\bigskip

\newpage

\section{Introduction and motivation}

\noindent Pattern recognition (or classification or discrimination) is about
predicting the unknown nature of an observation: an observation is a
collection of numerical measurements, represented by a vector $x$ belonging
to some measurable space $\mathcal{X}$. The unknown nature of the
observation is denoted by $y$ belonging to a measurable space $\mathcal{Y}$.
In pattern recognition, the goal is to create a measurable map $\phi :%
\mathcal{X}\rightarrow \mathcal{Y}$; $\phi (x)$ which represents one's
prediction of $y$ given $x$. The error of a prediction $\phi (x)$ when the
true value is $y$ is measured by $L(y,\phi (x))$, where the loss function $%
L\in \mathcal{Y}^{2}\rightarrow \mathbb{R}_{+}$. For simplicity, we suppose $%
L\leq 1$. In a probabilistic setting, the distribution
$\mathbb{P}$ of the random variable $(X,Y)\in \mathcal{X}\times
\mathcal{Y}$ describes the probability of encountering a
particular pair in practice. The performance of $\phi $, that is
how the predictor can predict future data, is measured by the
risk $R(\phi ):=\mathbb{E}_{(X,Y)}L(Y,\phi (X))$. In practice, we
have access to $n$ independent, identically distributed
($i.i.d.$) random pairs $(X_{i},Y_{i})_{1\leq i\leq n}$ sharing
the same distribution as $(X,Y) $ called the learning sample and
denoted $\mathcal{D}_{n}$. A learning algorithm $\Phi $ is
trained on the basis of $\mathcal{D}_{n}$. Thus,
$\Phi $ is a measurable map from $\mathcal{X}\times \cup _{n}(\mathcal{X}%
\times \mathcal{Y})^{n}$ to $\mathcal{Y}$. $Y$ is predicted by $\Phi (X,%
\mathcal{D}_{n}).$ The performance of $\Phi (.,\mathcal{D}_{n})$ is measured
by the conditional risk called the generalization error denoted by $%
\widetilde{R}_{n}:=\mathbb{E}_{(X,Y)}[L(Y,\Phi(X,\mathcal{D}_{n}))\mid
\mathcal{D}_{n}]$ with $(X,Y)\sim \mathbb{P}$ independent of
$\mathcal{D}_{n} $ and with the following equivalent notation for
the conditional expectation of $h(X,Y)$ given $Y$:
$\mathbb{E}_{X}h(X,Y)$. In the following, if there is no
ambiguity, we will also allow the notation $\phi
(X,\mathcal{D}_{n})$ instead of $\Phi (X,\mathcal{D}_{n})$.
Notice that $\widetilde{R}_{n}$ is a random variable measurable
with respect to $\mathcal{D}_{n}$.

\bigskip

\noindent An important question is: \textit{The distribution $\mathbb{P}$ of
the generating process being unknown, can we estimate how good a predictor
trained on a learning sample of size $n$ is? In other words, can we estimate
the generalization error $\widetilde{R}_{n}$?} This fundamental statistical
problem is referred to ''choice and assessment of statistical predictions'' %
\cite{STO74} . Many estimates have been proposed, among them the
resubstitution estimate (or training estimate). The predictor is
trained using the entire learning sample $\mathcal{D}_{n}$, and
an estimate of the prediction is obtained by running the same
learning process through the predictor and comparing predicted
and actual responses. Thus, the
resubstitution estimate $\widehat{R}_{n}:=\frac{1}{n}\sum_{i=1}^{n}L(Y_{i},%
\phi (X_{i},\mathcal{D}_{n}))$ can severely underestimate the
bias. It can even drop to zero for some machine learning even
though the generalization error is nonzero (for example, the
$1-$nearest neighbor). The difficulty arises from the fact that
the learning sample is used both for training and testing. In
order to get rid of this downward bias, the estimation of the
generalization error based on sample reuse have been favored among
practitioners. Quoting \cite{HTF01}: \textit{Probably the
simplest and most widely used method for estimating prediction
error is cross-validation}.
However, the role of cross-validation estimator, denoted by $\widehat{R}%
_{CV} $, is far from being well understood in a general setting. In
particular, the following problems remain partially solved: ''Is $\widehat{R}%
_{CV}$ a good estimator of the generalisation error?'', ''How should one
choose $k$ in a $k$-fold cross-validation'' or ''Does cross-validation
outperform the resubstitution error ?''. The purpose of this paper is to
give a partial answer to the first two questions.

\bigskip

\noindent We introduce our \textbf{main result} for symmetric
cross-validation procedures. We divide the learning sample into two samples:
the training sample and the test sample, to be defined below. We denote by $%
p_{n}$ the percentage of elements in the test sample such that
$np_n$ is an integer. For empirical risk
minimizers over a class of predictors with finite VC-dimension $V_{\mathcal{C%
}}$, to be defined below, we have the following concentration inequality,
for all $\varepsilon >0$:
\begin{equation*}
\Pr \mathbb{(}|\widehat{R}_{CV}-\widetilde{R}_{n}|\geq \varepsilon )\leq
B(n,p_{n},\varepsilon )+V(n,p_{n},\varepsilon ),
\end{equation*}

with

\begin{itemize}
\item $B(n,p_{n},\varepsilon )=\displaystyle5(2n(1-p_{n})+1)^{\frac{4V_{%
\mathcal{C}}}{1-p_{n}}}\exp (-\frac{n\varepsilon ^{2}}{64})$

\item $V(n,p_{n},\varepsilon )=\displaystyle\min \left( \exp (-\frac{%
np_{n}\varepsilon ^{2}}{2}),\frac{16}{\varepsilon }\sqrt{\frac{V_{\mathcal{C}%
}(\ln (2(1-p_{n})+1)+4)}{n(1-p_{n})}}\right) .$
\end{itemize}

\bigskip

\noindent The term $B(n,p_{n},\varepsilon )$ is a
Vapnik-Chernovenkis-type bound
controlled by the size of the training sample $n(1-p_{n})$ whereas the term $%
V(n,p_{n},\varepsilon )$ is the minimum between a Hoeffding-type
term controlled by the size of the test sample $np_{n}$, a
polynomial term controlled by the size of the training sample. As
the percentage of observations in the test sample $p_{n}$
increases, the $V(n,p_{n},\varepsilon )$ term decreases but the
$B(n,p_{n},\varepsilon ) $ term increases.

%Notice that this bound is worse than the Vapnik-like bound and
%thus can be called a ''sanity-check bound'' in the spirit of
%\cite{KR99}.

\noindent The difference from the previous results on estimation of $%
\widetilde{R}_{n}$ is in the following:

\begin{itemize}
\item our bounds for intensive cross-validation procedures (i.e. $k$-fold
cross-validation or leave-$\upsilon $-out cross-validation) are not worse
than those for hold-out cross-validation.

\item our inequalities not only depend on the percentage of observations in
the learning sample $p_{n}$ but also on the precise type of cross-validation
procedure: this is why we can discriminate between $k$-fold cross-validation
and hold-out cross-validation even if $p_{n}$ is the same.

\item we show that the size of the test sample does not need to grow to
infinity for the cross-validation procedure to be consistent for the
estimation of the generalization error.
\end{itemize}

\bigskip

\noindent Using these probability bounds, we can then deduce that the
expectation of the difference between the generalization error and the
cross-validation estimate $\mathbb{E}_{\mathcal{D}_{n}}|\widehat{R}_{CV}-%
\widetilde{R}_{n}|$ is of order%$\displaystyle
%O(\sqrt{ {V\ln(n(1-p_n))}{n(1-p_n)}}+\sqrt{\frac{1}{np_n}})$.
 $O_{n}(\sqrt{V_{\mathcal{C}}\ln (n(1-p_{n}))/n(1-p_{n})}+\sqrt{1/np_{n}})$.
As far as $\mathbb{E}_{\mathcal{D}_{n}}|\widehat{R}_{CV}-\widetilde{R}_{n}|$
is concerned, we can define a splitting rule: the percentage of elements $%
p_{n}$ in the test sample should be proportional to $\frac{1}{1+V_{\mathcal{C%
}}^{1/3}}$, i.e. the larger the class of predictors is, the smaller the test
sample in the cross-validation should be.\bigskip

\noindent The paper is organized as follows. In the next section, we give a
short review of literature. We detail the main cross-validation procedures
and we summarize the previous results for the estimation of generalization
error. In Section 3, we introduce the main notations and definitions.
Finally, in Section 4, we introduce our results, in terms of concentration
inequalities. In companion papers, we will show that in some cases, the
cross-validation estimate can outperform the training estimate and prove
that cross-validation can work out with infinite VC-dimension predictor.

\section{Short Review of the literature on cross-validation}

\noindent The cross-validation $\widehat{R}_{CV}$ includes
leave-one-out cross-validation, $k$-fold cross-validation,
hold-out cross-validation (or split sample), leave-$\upsilon
$-out cross-validation (or Monte Carlo cross-validation or
bootstrap cross-validation). In leave-one-out cross-validation, a
single sample of size $n$ is used. Each member of the sample in
turn is removed, the full modeling method is applied to the
remaining $n-1$ members, and the fitted model is applied to the
hold-backmember. An early (1968) application of this approach to
classification is that of \cite{LM68}. \cite{AL68} gave perhaps
the first application in multiple regression and \cite{GEI75}
sketches other applications. However, this special form of
cross-validation has well-known limitations, both theoretical and
practical, and a number of authors have considered more general
multifold cross-validation procedures \cite{BREI84}
; \cite{BREI92} ; \cite{BUR89} ; \cite{DGL96} ; \cite{GEI75} ; %
\cite{GYO02} ; \cite{McC76} ; \cite{PIC84} ; \cite{RIP96} ; %
\cite{SHAO93} ; \cite{ZHA93} ). The $k$-fold procedure divides
the learning sample into $k$ equally sized \ folds. Then, it
produces a predictor by training on $k-1$ folds and testing on
the remaining fold. This
is repeated for each fold, and the observed errors are averaged to form the $%
k$-fold estimate. Leave-$\upsilon $-out cross-validation is a more elaborate
and expensive version of cross-validation that involves leaving out all
possible subsamples of $\upsilon $ cases. In the split-sample method or
hold-out, only a single subsample (the training sample) is used to estimate
the generalization error, instead of $k$ different subsamples; i.e., there
is no crossing. Intuitively, there is a tradeoff between bias and variance
in cross-validation procedures. Typically, we expect the leave-one-out
cross-validation to have a low bias (the generalization error of a predictor
trained on $n-1$ pairs should be close to the generalization error of a
predictor trained on the $n$ pairs) but a high variance. Leave-one-out
cross-validation often works well for estimating generalization error for
continuous loss functions such as the squared loss, but it may perform
poorly for discontinuous loss functions such as the indicator loss. On the
contrary, $k$-fold cross-validation or leave-$\upsilon $-out
cross-validation are expected to have a higher bias but a smaller variance
due to resampling.

\bigskip

\noindent With the exception of \cite{BUR89}, theoretical
investigations of multifold cross-validation procedures have
first concentrated on linear models
(\cite{Li87};\cite{SHAO93};\cite{ZHA93}). Results of \cite{DGL96}
and \cite{GYO02} are discussed in Section 3. The first finite
sample results are due to \cite{DEWA79} and concern $k$-local
rules algorithms
under leave-one-out and hold-out cross-validation. More recently, %
\cite{HOL96, HOL96bis} derived finite sample results for the hold-out, $k-$%
fold and leave-one-out cross-validations for finite VC\ algorithms in the
realisable case (the generalization error is zero). But the bounds for $k-$%
fold cross-validation are $k$ times worse than for hold-out
cross-validation. \cite{BKL99} have emphasized when $k-$fold can
out perform hold-out cross-validation in a particular case of
$k$-fold predictor. \cite{KR99} has extended such results in the
case of stable algorithms for the leave-one-out cross-validation
procedure. \cite{KEA95} also derived results for hold-out
cross-validation for VC algorithms without the realisable
assumption. However, the bounds obtained are ''sanity check
bounds'' in the sense that they are not better than classical
Vapnik-Chernovenkis's bounds. \cite{DUD04BIS} derived finite
sample results for the distance between the cross-validation
estimate and a special benchmark and proved asymptotic results
for the relation between the cross-validation risk and the
generalization error. To our knowledge, bounds for intensive
cross-validation procedures are missing. This might be due to the
lack of
independence between the crossing terms of the cross-validated estimate %
\cite{KMNR95}.

\section{Notations and definitions}

\noindent We introduce here useful definitions to define the various
cross-validation procedures. First, we define binary vectors, i.e. $%
V_{n}=(V_{n,i})_{1\leq i\leq n}$ is a vector of size $n$, such that for all $%
i,V_{n,i}\in \{0,1\}$ and $\sum_{i}V_{n,i}\neq 0$. Consequently, knowing the
binary vector, we can define the subsample associated with it: $\mathcal{D}%
_{V_{n}}:=\{(X_{i},Y_{i})\in \mathcal{D}_{n}|V_{n,i}=1,1\leq i\leq n\}$. The
weighted empirical error of $\varphi $ is denoted by $\hat{R}%
_{V_{n}}(\phi )$ and defined by:
\begin{equation*}
\hat{R}_{V_{n}}(\phi ):=\frac{1}{\sum_{i=1}^{n}V_{n,i}}%
\sum_{i=1}^{n}V_{n,i}L(Y_{i},\phi (X_{i})).
\end{equation*}%
\noindent For $\hat{R}_{1_{n}}$, with $1_{n}$ the binary vector of size $n$
with $1$ at every coordinate, we will use the simpler notation $\hat{R}_{n}$%
. For a predictor trained on a subsample, we define:
\begin{equation*}
\phi _{V_{n}}(.):=\Phi (.,\mathcal{D}_{V_{n}}).
\end{equation*}

%\begin{definition}[Weighted empirical measure] For a given non-negative vector
%$(V_{n,i}%
%)_{1 \leq i \leq n}%
%$ of size $n$, we define a weighted empirical measure:
%$$ \mathbb{P}_{n,V_n}
%:= \frac{1}{\sum_i V_{n,i}}
%\sum_{i=1}^{n} V_{n,i} \delta_{(X_{i},Y_{i}%
%)}$$
%with $\delta_{(X_{i},Y_{i})}$ the Dirac measure at $(X_{i}%
%,Y_{i})$
%\end{definition}

\noindent With the previous notations, notice that the predictor trained on
the learning sample $\phi (.,\mathcal{D}_{n})\ $can be denoted by $%
\phi _{1_{n}}(.)$. We will allow the simpler notation $\phi
_{n}(.)$. The learning sample is divided into two disjoint
samples: the training
sample of size $n(1-p_{n})$ and the test sample of size $np_{n}$, where $%
p_{n}$ is the percentage of elements in the test sample. To represent the
training sample, we define a random binary vector $V_{n}^{tr}$ of size $n$
independent of $\mathcal{D}_{n}$. $V_{n}^{tr}$ is called the training
vector. We define the test vector by $V_{n}^{ts}:=1_{n}-V_{n}^{tr}$ to
represent the test sample.

\noindent The distribution of $V_{n}^{tr}$ characterizes all the
cross-validation procedures described in the previous section. Using our
notations, we can now define the cross-validation estimator.

\begin{definition}[Cross-validation estimator]
With the previous notations, the generalized cross-validation error of $%
\phi _{n}$ denoted by $\widehat{R}_{CV}$ is defined by the
conditionnal expectation of $\hat{R}_{V_{n}^{ts}}(\phi
_{V_{n}^{tr}})$ with respect to the random vector $V_{n}^{tr}$
given $\mathcal{D}_{n}$:
\begin{equation*}
\widehat{R}_{CV}:=\mathbb{E}_{V_{n}^{tr}}\hat{R}_{V_{n}^{ts}}(\phi
_{V_{n}^{tr}}).
\end{equation*}
\end{definition}

\noindent We will give here some examples of distributions of $V_{n}^{tr}$
to show that we retrieve cross-validation procedures described previously.
Suppose $n/k$ is a integer.\ The $k$-fold procedure divides the data into $k$
equally sized \ folds. It then produces a predictor by training on $k$-1
folds and testing on the remaining fold. This is repeated for each fold, and
the observed errors are averaged to form the $k$-fold estimate.

\begin{example}[$k$-fold cross-validation]
\begin{align*}
\Pr(V_{n}^{tr} & =(\underbrace{0,\ldots,0}_{n/k\text{ observations}},%
\underbrace{1,\ldots,1}_{n(1-1/k)\text{ observations}}))=\frac{1}{k} ,\\
\Pr(V_{n}^{tr} & =(\underbrace{1,\ldots,1}_{n/k\text{ observations}},%
\underbrace{0,\ldots,0}_{n/k\text{ observations}},\underbrace{1,\ldots ,1}%
_{n(1-2/k)\text{ observations}}))=\frac{1}{k} ,\\
& \ldots \\
\Pr(V_{n}^{tr} & =(\underbrace{1,\ldots,1}_{n(1-1/k)\text{
observations}},\underbrace{0,\ldots,0}_{n/k\text{ observations}}))=\frac{1}{k%
}.
\end{align*}
\end{example}

\noindent We provide another popular example: the leave-one-out
cross-validation. In leave-one-out cross-validation, a single sample of size
$n$ is used. Each member of the sample in turn is removed, the full modeling
method is applied to the remaining $n-1$ members, and the fitted model is
applied to the hold-backmember.

	\begin{example}[leave-one-out cross-validation]
	\begin{align*}
	\Pr(V_{n}^{tr}& =(0,1,\ldots ,1))=\frac{1}{n} \\
	\Pr(V_{n}^{tr}& =(1,0,1,\ldots ,1))=\frac{1}{n} \\
	& \ldots  \\
	\Pr(V_{n}^{tr}& =(1,\ldots ,1,0))=\frac{1}{n}.
	\end{align*}
	\end{example}
	
\noindent We denote by $R_{opt}$ the minimal generalization error attained
among the class of predictors $\mathcal{C}$, $R_{opt}=\inf_{\phi \in
\mathcal{C}}R(\phi )$. In the sequel, we suppose that $\phi _{n}$ is an
empirical risk minimizer over the class $\mathcal{C}$. For simplicity, we
suppose the infimum is attained i.e. $\phi _{n}=\arg \min_{\phi \in \mathcal{%
C}}\widehat{R}_{n}(\phi )$. Notice that $R_{opt}$ is a parameter of the
unknown distribution $\mathbb{P}_{(X,Y)}$ whereas $\widetilde{R}_{n}$ is a
random variable.

\bigskip

\noindent At last, recall the definitions of:

\begin{definition}[Shatter coefficients]
Let $\mathcal{A}$ be a collection of measurable sets. For $(z_{1,\ldots
,}z_{n})$ $\in \{\mathbb{R}^{d}\}^{n}$, let $N_{\mathcal{A}}(z_{1,\ldots
,}z_{n})$ be the number of differents sets in
\begin{equation*}
\{\{z_{1},\ldots ,z_{n}\}\cap A;A\in \mathcal{A}\}.
\end{equation*}

The n-shatter coefficient of $\mathcal{A}$ is
\begin{equation*}
\mathcal{S}(n,\mathcal{A})=\max_{(z_{1,\ldots ,}z_{n})\in \{\mathbb{R}%
^{d}\}^{n}}N_{\mathcal{A}}(z_{1,\ldots ,}z_{n}).
\end{equation*}

That is, the shatter coefficient is the maximal number of different subsets
of $n$ points that can be picked out by the class of sets $\mathcal{A}$.
\end{definition}

\begin{definition}[VC dimension]
Let $\mathcal{A}$ be a collection of sets with $\mathcal{A}\geq
2.$ The largest integer $k\geq 1$ for which
$\mathcal{S}(k,\mathcal{A})=2^{k}$ is denoted by
$V_{\mathcal{C}}$, and it is called the Vapnik-Chernovenkis
dimension (or VC dimension) of the class $\mathcal{A}$. If
$\mathcal{S}(n,
\mathcal{A})=2^{n}$ for all n, then by definition $V_{\mathcal{C}%
}=\infty .$
\end{definition}

\noindent A class of predictors $\mathcal{C}$ is said to have a finite
VC-dimension $V_{\mathcal{C}}$ if the dimension of the collection of sets $%
\{A_{\phi ,t}:\phi \in \mathcal{C},t\in \lbrack 0,1]\}$ is equal to $%
V_{\mathcal{C}}$, where $A_{\phi ,t}=\{(x,y)/L(y,\phi (x))>t\}$.

\section{Results}

\label{results}

\subsection{Hypotheses $\mathcal{H}$}

\noindent In the sequel, we suppose that the training sample and
the test sample are disjoint and that the number of observations
in the training sample and in the test sample are respectively
$n(1-p_{n})$ and $np_{n}$. Moreover, we suppose also that the
$\phi _{n}$ is an empirical risk minimizer on a sample with
finite VC-dimension $V_{\mathcal{C}}$ and $L$ a loss function
bounded by $1$. We also suppose that the predictors are symmetric
according to the training sample, i.e. the predictor does not
depend on the order of the observations in $\mathcal{D}_{n}$.
Eventually, the cross-validation are symmetric i.e. $\Pr
(V_{n,i}^{tr}=1)$ does not depend on $i$, this excludes the
hold-out cross-validation. \textbf{We denote these hypotheses by
$\mathcal{H}$.}

\bigskip

\noindent %$\mathbb{P}$ stands for $\mathbb{P}_{%
%\mathcal{D}_{n}}$.
We will show upper bounds of the kind $\Pr (|\widehat{R}_{CV}-\widetilde{R}%
_{n}|\geq \varepsilon )\leq B(n,p_{n},\varepsilon
)+V(n,p_{n},\varepsilon )$ with $\varepsilon >0$. The term
$B(n,p_{n},\varepsilon )$ is a Vapnik-Chernovenkis-type bound
whereas the term $V(n,p_{n},\varepsilon )$ is a Hoeffding-like
term controlled by the size of the test sample $np_{n}$. This
bound gives can be interpreted as a quantitative answer to the
bias-variance trade-off question. As the percentage of
observations in the test sample $p_{n}$
increases, the $V(n,p_{n},\varepsilon )$ term decreases but the $%
B(n,p_{n},\varepsilon )$ term increases. Notice that this bound
is worse than the Vapnik-Chernovenkis-type bound and thus can be
called a ''sanity-check bound'' in the spirit of \cite{KR99}.
Even though these bounds are valid for almost all the
cross-validation procedures, their relevance depends highly on
the percentage $p_{n}$ of elements in the test sample; this is
why we first classify them according to $p_{n}$. At last, notice
that our bounds can be refined using chaining arguments. However,
this is not the purpose of this paper.

\subsection{Cross-validation with large test samples}

The first result deals with large test samples, i.e. the bounds are all the
better if $np_{n}$ is large. Note that this result excludes the hold-out
cross-validation because it does not make a symmetric use of the data.

\begin{proposition}[Large test sample]
\label{largets1}%
Suppose that $\mathcal{H}$ holds. Then, we have for all $\varepsilon>0$,
$$\Pr(\widehat{R}_{CV}-\widetilde{R}%
_{n}\geq\varepsilon )\leq B(n,p_n,\varepsilon)
+V(n,p_n,\varepsilon),$$

with
\begin{itemize}
\item $B(n,p_n,\varepsilon)=\displaystyle 4(2 n (1-p_n)+1)^{\frac{4V_{\mathcal{C}}}{1-p_{n}%
}}\exp(-\frac{n \epsilon^{2}}{25}),$
\item $V(n,p_n,\varepsilon)=\displaystyle \exp(-\frac{2np_{n}%
\varepsilon^{2}}%
{25}).$
\end{itemize}
\end{proposition}

\bigskip

\noindent First, we begin with a useful lemma( for the proof, see Appendices)

\begin{lemma}
\label{ch1:lemme1}

Under the assumption of Proposition \ref{largets1}, we have for all $%
\varepsilon >0,$%
\begin{equation*}
\Pr \mathbb{(E}_{V_{n}^{tr}}\sup_{\phi \in \mathcal{C}}(\widehat{R}%
_{V_{n}^{tr}}(\phi )-R(\phi ))\geq \varepsilon )\leq (\mathcal{S}%
(2n(1-p_{n}),\mathcal{C}))^{\frac{4}{1-p_{n}}}e^{-n\varepsilon ^{2}},
\end{equation*}%
and symmetrically

\begin{equation*}
\Pr \mathbb{(E}_{V_{n}^{tr}}\sup_{\phi \in \mathcal{C}}(R(\phi )-\widehat{R}%
_{V_{n}^{tr}}(\phi ))\geq \varepsilon )\leq (\mathcal{S}(2n(1-p_{n}),%
\mathcal{C}))^{\frac{4}{1-p_{n}}}e^{-n\varepsilon ^{2}}.
\end{equation*}
\end{lemma}

\bigskip

\noindent {\bfseries  Proof of proposition \ref{largets1}.}

\noindent Recall that $\phi _{n}$ is based on empirical risk minimization.
Moreover, for simplicity, we have supposed the infimum is attained \textit{%
i.e.} $\phi _{n}=\arg \min_{\phi \in \mathcal{C}}\widehat{R}_{n}(\phi )$.
Define $\bar{R}_{n(1-p)}:=\mathbb{E}_{V_{n}^{tr}}R(\phi _{V_{n}^{tr}})$.

\bigskip \noindent We have by splitting according to $\bar{R}_{n(1-p)}$:
\begin{equation*}
\Pr \mathbb{(}\widehat{R}_{CV}-\widetilde{R}_{n}\geq 5\varepsilon )\leq
\underbrace{\Pr \mathbb{(}\widehat{R}_{CV}-\bar{R}_{n(1-p)}\geq \varepsilon )%
}_{V}+\underbrace{\Pr( \bar{R}_{n(1-p)}-\widetilde{R}_{n}\geq
4\varepsilon )}_{B}.
\end{equation*}

\noindent Notice that $\mathbb{E}_{\mathcal{D}_{n}}(\widehat{R}_{CV}-\bar{R}%
_{n(1-p)})=0$. Intuitively, $V$ corresponds to the variance term and is
controlled in some way by the resampling plan. On the contrary, in the general setting, $\mathbb{E}_{%
\mathcal{D}_{n}}(\bar{R}_{n(1-p)}-\widetilde{R}_{n})\neq 0$, and
$B$ is the bias term and measures the discrepancy between the
error rate of size $n$ and of size $n(1-p_{n}).$

\noindent The first term $V$ can be bounded via Hoeffding's
inequality, as follows

\begin{align*}
V& =\Pr (\mathbb{E}_{V_{n}^{tr}}(\widehat{R}_{V_{n}^{ts}}(\phi
_{V_{n}^{tr}})-R(\phi _{V_{n}^{tr}}))\geq \varepsilon ) \\
& \leq \inf_{s>0}e^{-s\varepsilon }\mathbb{E}e^{s\mathbb{E}_{V_{n}^{tr}}(%
\widehat{R}_{V_{n}^{ts}}(\phi _{V_{n}^{tr}})-R(\phi _{V_{n}^{tr}}))}\text{
(by Chernoff's bound).}
\end{align*}%
\noindent Then, by Jensen's inequality, we have
\begin{equation*}
V\leq \inf_{s>0}e^{-s\varepsilon }\mathbb{E_{\mathcal{D}_n}} \mathbb{E}_{V_{n}^{tr}}e^{s(\widehat{R}%
_{V_{n}^{ts}}(\phi _{V_{n}^{tr}})-R(\phi _{V_{n}^{tr}}))}.
\end{equation*}%
\noindent Thus, for $\mathbf{v_{n}^{tr},v_{n}^{ts}}$ fixed
vectors, we have by linearity of expectation and the i.i.d
assumption

\begin{align*}
V& \leq \inf_{s>0}e^{-s\varepsilon }\mathbb{E}e^{s(\widehat{R}_{\mathbf{%
v_{n}^{ts}}}(\phi _{\mathbf{v_{n}^{tr}}})-R(\phi _{\mathbf{v_{n}^{tr}}}))} \\
& \leq \inf_{s>0}e^{-s\varepsilon }\mathbb{E}_{\mathcal{D}_{\mathbf{v}_{n}^{tr}}}\mathbb{E}(e^{s(\widehat{R}_{\mathbf{v}%
_{n}^{ts}}(\phi _{\mathbf{v}_{n}^{ts}})-R(\phi _{\mathbf{v}_{n}^{tr}}))}\mid
\mathcal{D}_{\mathbf{v}_{n}^{tr}}).
\end{align*}%
\noindent

\noindent Finally, by lemma 1 in \cite{Lug03} since $\mathbb{E}(\widehat{R}_{%
\mathbf{v}_{n}^{ts}}(\phi _{\mathbf{v}_{n}^{ts}})-R(\phi _{\mathbf{v}%
_{n}^{tr}})\mid \mathcal{D}_{\mathbf{v}_{n}^{tr}})=0$ and the conditional
independence:

\begin{equation*}
V\leq \inf_{s>0}e^{-s\varepsilon }\mathbb{E}e^{\frac{s^{2}}{8np_{n}}}\leq
e^{-2np_{n}\varepsilon ^{2}}.
\end{equation*}

\noindent The second term may be treated by introducing the optimal error $%
R_{opt}$ which should be close to $\widetilde{R}_{n}$,

\begin{align}
B& =\Pr \mathbb{(}\bar{R}_{n(1-p)}-\widetilde{R}_{n}\geq 4\varepsilon )
\notag \\
& =\Pr \mathbb{(E}_{V_{n}^{tr}}(R(\phi _{V_{n}^{tr}})-\widehat{R}%
_{_{V_{n}^{tr}}}(\phi _{V_{n}^{tr}})+\widehat{R}_{_{V_{n}^{tr}}}(\phi
_{V_{n}^{tr}})-R_{opt})+R_{opt}-\widetilde{R}_{n}\geq 4\varepsilon ).  \notag
\end{align}

\noindent Using the supremum and the fact that $\phi _{V_{n}^{tr}}$ is an
empirical risk minimizer, we obtain:
\begin{align*}
B& \leq \Pr \mathbb{(E}_{V_{n}^{tr}}\sup_{\phi \in \mathcal{C}}(R(\phi )-%
\widehat{R}_{V_{n}^{tr}}(\phi ))+\mathbb{E}_{V_{n}^{tr}}\inf_{\phi \in
\mathcal{C}}\widehat{R}_{V_{n}^{tr}}(\phi )-\inf_{\phi \in \mathcal{C}%
}R(\phi ) \\
& \quad +R_{opt}-\hat{R}_{n}+\hat{R}_{n}-\widetilde{R}_{n}\geq 4\varepsilon
).
\end{align*}

\noindent Then, since $\inf (A)-\inf (B)\leq \sup (A-B)$ and by
definition of $\phi _{n}$, we deduce
\begin{align}
B& \leq \Pr \mathbb{(E}_{V_{n}^{tr}}\sup_{\phi \in \mathcal{C}}(R(\phi )-%
\widehat{R}_{V_{n}^{tr}}(\phi ))\geq \varepsilon )+\Pr \mathbb{(E}%
_{V_{n}^{tr}}(\sup_{\phi \in \mathcal{C}}(\widehat{R}_{V_{n}^{tr}}(\phi
)-R(\phi ))\geq \varepsilon )  \notag \\
& \quad +\Pr \mathbb{(}\sup_{\phi \in \mathcal{C}}(R(\phi )-\widehat{R}%
_{n}(\phi ))\geq \varepsilon )+\Pr \mathbb{(}\sup_{\phi \in \mathcal{C}}(%
\widehat{R}_{n}(\phi )-R(\phi ))\geq \varepsilon ).  \notag
\end{align}

\noindent Thus, by Lemma \ref{ch1:lemme1}, we get
\begin{equation*}
B\leq 2(\mathcal{S}(2n(1-p_{n}),\mathcal{C}))^{\frac{4}{1-p_{n}}%
}e^{-n\varepsilon ^{2}}+2\mathcal{S}(2n,\mathcal{C})^{4}e^{-n\varepsilon
^{2}}.
\end{equation*}

\noindent Recall the following result (see e.g. \cite{DGL96})
\begin{equation}
\forall n,\mathcal{S}(n,\mathcal{C})\leq (n+1)^{V_{\mathcal{C}}}.
\label{eq:shatter}
\end{equation}%
\noindent Thus, we finally obtain

\begin{align}
B& \leq 2(2n(1-p_{n})+1)^{\frac{4V_{\mathcal{C}}}{1-p_{n}}}e^{-n\varepsilon
^{2}}+2(2n+1)^{4V_{\mathcal{C}}}e^{-n\varepsilon ^{2}}  \notag \\
& \leq 4(2n(1-p_{n})+1)^{\frac{4V_{\mathcal{C}}}{1-p_{n}}}e^{-n\varepsilon
^{2}}.  \notag
\end{align}

\bigskip$\Box$

\noindent Next, we obtain
\begin{proposition}[Large test sample]
\label{largets}%
Suppose that $\mathcal{H}$ holds. Then, we have, for all $\varepsilon>0$,
$$\Pr(\widetilde{R}%
_{n} - \widehat{R}_{CV} \geq\varepsilon )\leq (2 n +1)^{4V}
\exp(-n \varepsilon^{2}).$$
\end{proposition}

\noindent {\bfseries  Proof }

\noindent First, the following lemma holds (for the proof, see appendices),

\begin{lemma}
\label{lemme1}

Suppose that $\mathcal{H}$ holds, then we have $\widehat{R}%
_{CV}\geq \widehat{R}_{n}.$
\end{lemma}

\noindent Thus,

\begin{equation*}
\Pr \mathbb{(}\widetilde{R}_{n}-\widehat{R}_{CV}\geq \varepsilon )\leq \Pr
\mathbb{(}\widetilde{R}_{n}-\widehat{R}_{n}\geq \varepsilon )\leq \mathcal{S}%
(2n,\mathcal{C})^{4}e^{-\varepsilon ^{2}n}\leq (2n+1)^{4V_{\mathcal{C}%
}}e^{-n\varepsilon ^{2}}.
\end{equation*}

\bigskip$\Box$

\noindent Using the two previous results, we have a concentration inequality
for the absolute error $|\widehat{R}_{CV}-\widetilde{R}_{n}|$,

\begin{corollary}[Absolute error for large test sample]
Suppose that $\mathcal{H}$ holds. Then, we have, for all $\varepsilon >0$,
\begin{equation*}
\Pr (|\widetilde{R}_{n}-\widehat{R}_{CV}|\geq \varepsilon )\leq
B(n,p_{n},\varepsilon )+V(n,p_{n},\varepsilon ),
\end{equation*}

with

\begin{itemize}
\item $B(n,p_{n},\varepsilon )=\displaystyle5(2n(1-p_{n})+1)^{\frac{4V_{%
\mathcal{C}}}{1-p_{n}}}\exp (-\frac{n\varepsilon ^{2}}{25}),$

\item $V(n,p_n,\varepsilon)=\displaystyle \exp(-\frac{2np_{n}\varepsilon^{2}%
}{25}).$
\end{itemize}
\end{corollary}

\noindent With the previous concentration inequality, we can bound from
above the expectation of $|\widetilde{R}_{n}-\widehat{R}_{CV}|$:

\begin{corollary}[$L_{1}$ error for large test sample]
\label{cor:l1-largets}Suppose that $\mathcal{H}$ holds. Then, we have,
\begin{equation*}
\mathbb{E}|\widehat{R}_{CV}-\widetilde{R}_{n}|\leq 10\sqrt{\frac{V(\ln
(2n(1-p_{n})+1)+4)}{n(1-p_{n})}}+5\sqrt{\frac{2}{np_{n}}}.
\end{equation*}
\end{corollary}

\noindent {\bfseries  Proof. }

\noindent This is a direct consequence of the following lemma:

\begin{lemma}[\cite{DGL96}]
\label{ch1:lem esperance} Let $X$ be a nonnegative random
variable. Let $K,C$
nonnegative real such that $C\geq1$. Suppose that for all $\varepsilon>0$ $%
\mathbb{P}(X \geq \varepsilon)\leq C\exp(-K \varepsilon^2)$. Then:
\begin{equation*}
\mathbb{E}X \leq \sqrt{\frac{\ln(C)+2}{K}}.
\end{equation*}
\end{lemma}

\bigskip$\Box$

\subsection{Cross-validation with small test samples}

The previous bound is not relevant for all small test samples (typically
leave-one-out cross-validation) since we are not assured that the variance
term converges to $0$ (in leave-one-out cross-validation, $%
V(n,p_{n},\varepsilon )=\displaystyle\exp (-2\varepsilon
^{2}/25)$). However, under $\mathcal{H}$, cross-validation with
small test samples works also, as stated in the next proposition.

\begin{proposition}[Small test sample]
\label{smallts}%
Suppose that $\mathcal{H}$ holds. Then, we have, for all $\varepsilon>0$,
\begin{align*}
\Pr(\widehat{R}%
_{CV}-\widetilde{R}_{n}
 \geq\varepsilon
)\leq &  B(n,p_n,\varepsilon) +V(n,p_n,\varepsilon),
\\
& \end{align*}%

with

\begin{itemize}
\item $B(n,p_n,\varepsilon)=\displaystyle 4(2 n (1-p_n)+1)^{\frac{4V_{\mathcal{C}}}{1-p_{n}%
}}\exp(-\frac{n \varepsilon^{2}}{64}),$
\item $V(n,p_n,\varepsilon)=\displaystyle \frac{1}{16 \varepsilon}\left(\sqrt{\frac{V_{\mathcal{C}}(\ln(2n(1-p_n)+1)+4)}{n(1-p_n)}}\right).$
\end{itemize}
\end{proposition}

\noindent For small test samples, we get the same conclusion but the rate of
convergence for the term $V$ is slower than for large test samples:
typically $O_{n}\left( \frac{1}{\varepsilon }\sqrt{\frac{\ln (n(1-p_{n}))}{%
n(1-p_{n})}}\right) $ against $O_{n}\left( \exp (-np_{n}\varepsilon
^{2})/8\right) .$

\bigskip

\noindent {\bfseries  Proof. }

\noindent Now, we get by splitting according to $\bar{R}_{n(1-p)}$:
\begin{equation*}
\Pr \mathbb{(}\widehat{R}_{CV}-\widetilde{R}_{n}\geq 8\varepsilon )\leq
\underbrace{\Pr \mathbb{(}\widehat{R}_{CV}-\bar{R}_{n(1-p)}\geq 4\varepsilon
)}_{V}+\underbrace{\Pr \mathbb{(}\bar{R}_{n(1-p)}-\widetilde{R}_{n}\geq
4\varepsilon )}_{B}.
\end{equation*}

\noindent First, from the proof of proposition \ref{largets}, we have $B\leq
4(2n(1-p_{n})+1)^{\frac{4V_{\mathcal{C}}}{1-p_{n}}}e^{-n\varepsilon ^{2}}.$
\bigskip

\noindent Secondly, notice that $\mathbb{E(}\widehat{R}_{CV}-\bar{R}%
_{n(1-p)})=0$. To control $V$, we will need the following lemma (for the
proof see appendices) which says that if a bounded random variable $X$ is
centered and is nonpositive with small probability then it is nonnegative
with also small probability.

\begin{lemma}
\label{ch1:monlemme}

If $|X|\leq1$ and $\mathbb{E}X=0$. Then for all $%
\varepsilon>0,$ we get
\begin{equation*}
\mathbb{P(}X\geq\varepsilon)\leq\frac{\int_{0}^{1}\mathbb{P(}X\leq -x)dx}{%
\varepsilon}.
\end{equation*}
\end{lemma}

\bigskip

\noindent Moreover, we have since $\widehat{R}_{CV}
\geq\widehat{R}_{n}$ by lemma \ref{lemme1}

\begin{align*}
\Pr \mathbb{(}\widehat{R}_{CV}-\bar{R}_{n(1-p)}\leq -4\varepsilon )& \leq
\Pr \mathbb{(}\widehat{R}_{n}-\bar{R}_{n(1-p)}\leq -4\varepsilon ) \\
& \leq \Pr \mathbb{(}\widehat{R}_{n}-\widetilde{R}_{n}\leq -\varepsilon
)+\Pr \mathbb{(}\widetilde{R}_{n}-\bar{R}_{n(1-p)}\leq -3\varepsilon ).
\end{align*}

\noindent Using lemma \ref{ch1:lemme1}, it follows:

\begin{align*}
\Pr \mathbb{(}\widehat{R}_{CV}-\bar{R}_{n(1-p)}\leq -4\varepsilon )& \leq
\mathcal{S}(2n,\mathcal{C})^{4}e^{-\varepsilon ^{2}n}+3\mathcal{S}%
(2n(1-p_{n}),\mathcal{C})^{\frac{4V_{\mathcal{C}}}{1-p_{n}}}e^{-n\varepsilon
^{2}} \\
& \leq 4(2n(1-p_{n})+1)^{\frac{4V_{\mathcal{C}}}{1-p_{n}}}e^{-n\varepsilon
^{2}}.
\end{align*}

\noindent Applying lemmas \ref{ch1:monlemme} and inequality
\ref{eq:shatter} allows to conclude.

$\Box$

\bigskip

\noindent We have the following complementary but not symmetrical result:

\begin{proposition}[Small test sample bis]
\label{smallts}%
Suppose that $\mathcal{H}$ holds. Then, we have for all $\varepsilon>0$,
$$
\mathbb{P}(\widetilde{R}_{n}-\widehat{R}%
_{CV}
 \geq\varepsilon
) \leq (2n+1)^{4V_{\mathcal{C}}} \exp(-n\varepsilon^2).
$$
\end{proposition}

\bigskip \noindent {\bfseries  Proof.}

\noindent We have since $\widehat{R}_{CV} \geq\widehat{R}_{n}$:

\begin{equation*}
\Pr \mathbb{(}\widetilde{R}_{n}-\widehat{R}_{CV}\geq \varepsilon )\leq \Pr
\mathbb{(}\widetilde{R}_{n}-\widehat{R}_{n}\geq \varepsilon )\leq \mathcal{S}%
(2n,\mathcal{C})^{4}e^{-\varepsilon ^{2}n}\leq (2n+1)^{4V_{\mathcal{C}%
}}e^{-n\varepsilon ^{2}}.
\end{equation*}

$\Box$

\noindent From this result, we deduce that,

\begin{corollary}[Absolute error for small test sample ]
Suppose that $\mathcal{H}$ holds. Then, we have for all $\varepsilon >0$,
\begin{equation*}
\Pr (|\widetilde{R}_{n}-\widehat{R}_{CV}|\geq \varepsilon )\leq
B(n,p_{n},\varepsilon )+V(n,p_{n},\varepsilon ),
\end{equation*}

\begin{itemize}
\item $B(n,p_{n},\varepsilon )=\displaystyle5(2n(1-p_{n})+1)^{\frac{4V_{%
\mathcal{C}}}{1-p_{n}}}\exp (-\frac{n\varepsilon ^{2}}{64})$

\item $V(n,p_{n},\varepsilon )=\displaystyle\frac{16}{\varepsilon }\sqrt{%
\frac{V_{\mathcal{C}}(\ln (2n(1-p_{n})+1)+4)}{n(1-p_{n})}}.$
\end{itemize}
\end{corollary}

\noindent Eventually, we get

\begin{corollary}[$L_{1}$ error for small test sample]
\label{largets}Suppose that $\mathcal{H}$ holds. Then, we have:
\begin{equation*}
\mathbb{E}|\widehat{R}_{CV}-\widetilde{R}_{n}|\leq 16\sqrt{\frac{V_{\mathcal{%
C}}\ln (2n(1-p_{n})+1)+4}{n(1-p_{n})}}\left( \ln \left( \sqrt{\frac{%
n(1-p_{n})}{V_{\mathcal{C}}(\ln (2n(1-p_{n})+1)+4)}}\right) +2\right) .
\end{equation*}
\end{corollary}

\noindent \bigskip{\bfseries  Proof. }

\noindent We just need lemma \ref{ch1:lem esperance} and the
following simple lemma

\begin{lemma}
Let $X$ a nonnegative random variable bounded by $1$, $A>0$ a real such that $\mathbb{P}%
(X\geq \varepsilon)\leq \frac{A}{\varepsilon}$, for all $\varepsilon>0$.
Then,
\begin{equation*}
\mathbb{E}(X) \leq A(1-\ln(A))
\end{equation*}
\end{lemma}

\bigskip$\Box$

\noindent Eventually, collecting the previous results, we can summarize the
previous results for upper bounds in probability with the following theorem:

\begin{theorem}[Absolute error for cross-validation]
\label{thm:sym}

\noindent Suppose that $\mathcal{H}$ holds. Then, we have for all $%
\varepsilon >0$,
\begin{equation*}
\Pr (|\widetilde{R}_{n}-\widehat{R}_{CV}|\geq \varepsilon )\leq
B_{sym}(n,p_{n},\varepsilon )+V_{sym}(n,p_{n},\varepsilon ),
\end{equation*}

with

\begin{itemize}
\item $B_{sym}(n,p_{n},\varepsilon )=\displaystyle5(2n(1-p_{n})+1)^{\frac{%
4V_{\mathcal{C}}}{1-p_{n}}}\exp (-\frac{n\varepsilon ^{2}}{64})$

\item $V_{sym}(n,p_{n},\varepsilon )=\displaystyle\min \left( \exp (-\frac{%
2np_{n}\varepsilon ^{2}}{25}),\frac{16}{\varepsilon }\sqrt{\frac{V_{\mathcal{C}%
}(\ln (2(1-p_{n})+1)+4)}{n(1-p_{n})}}\right) .$
\end{itemize}
\end{theorem}

\noindent An interesting consequence of this proposition is that the size of
the test is not required to grow to infinity for the consistency of the
cross-validation procedure in terms of convergence in probability.

%\bigskip
%\noindent Eventually, we have the following upper bound for the $L_1$ error of symmetric cross-validation procedures.

%\begin{thm}[$L_1$ error for symmetric cross-validation]
%\label{thm:sym}

%\noindent Suppose $\mathcal{H}$. Then, we have,
%\begin{small}
%\begin{align*}
%\mathbb{E}(|\widetilde{R}_{n}-\widehat{R}%
%_{CV}| \geq \varepsilon) \leq  & \min\left(10\sqrt{\frac{V(\ln(2n(1-p_n)+1)+4)}{n(1-p_n)}}+
%5\sqrt{\frac{2}{np_n}} , \\
%& \quad 16 \sqrt{\frac{V\ln(2n(1-p_n)+1)+4}{n(1-p_n)}}\left(\ln \left(
%\sqrt{\frac{n(1-p_n)}{V(\ln(2n(1-p_n)+1)+4)}} \right)+2\right) \right)
%\end{align*}
%\end{small}
%\end{thm}

\subsection{$k$-fold cross-validation}

\noindent For $k$-fold cross-validation, we can simply use the previous
bounds together. Thus, we get

\begin{proposition}[k-fold]
\label{k-fold}
 \noindent Suppose that $\mathcal{H}$ holds. Then, we have for all $\varepsilon>0$,
\begin{align*}
\Pr(|\widetilde{R}_{n}-\widehat{R}%
_{CV}| \geq \varepsilon) \leq  &B_{k}(n,p_n,\varepsilon)+
V_{k}(n,p_n,\varepsilon)
\end{align*}

with
\begin{itemize}
\item $B_{k}(n,p_n,\varepsilon)=\displaystyle 5(2 n (1-1/k)+1)^{\frac{4V_{\mathcal{C}}}{1-1/k%
}}\exp(-\frac{n \varepsilon^{2}}{64}) $
\item $V_{k}(n,p_n,\varepsilon)= \displaystyle \min \left( \exp(-\frac{2n
\varepsilon^2}{25k}),\frac{16}{\varepsilon}\sqrt{\frac{V_{\mathcal{C}}
(\ln(2(1-1/k)+1)+4)}{n(1-1/k)}}\right).$
\end{itemize}

\end{proposition}

\noindent Since $k \geq 2$, notice the previous bound can itself
be bounded by

\begin{equation*}
5(2n+1)^{8V_{\mathcal{C}}}\exp (-\frac{n\epsilon ^{2}}{64})+\min \left(
2\exp (-\frac{2n\varepsilon ^{2}}{25k}),\frac{16}{\varepsilon }\sqrt{\frac{%
(V_{\mathcal{C}}\ln (2n+1)+4)}{n}}\right) .
\end{equation*}

\bigskip

\noindent In fact, the bound for the variance term $(V)$ can be
improved by
averaging the $k$ training errors. This step emphasizes the interest of $k$%
-fold cross-validation against simpler cross-validation.

\begin{proposition}[k-fold]
\label{k-fold} Suppose that $\mathcal{H}$ holds. Then, in the
case of the $k$-fold cross-validation procedure, we have for all
$\varepsilon>0$:
$$
\Pr(\widehat{R}_{CV}-\hat{R}_{n(1-p_n)}
 \geq\varepsilon
)\leq 2^{\frac{1}{p_{n}}}\exp\left(-\frac{n\epsilon^{2}%
}{64(\sqrt{V_{\mathcal{C}}\ln(2(2np_n+1))}+2)}\right).\\
$$
\end{proposition}

\noindent Thus, averaging the observed errors to form the $k$-fold estimate
improves the term $V_{\mathcal{C}}$ from
\begin{equation*}
\displaystyle\min (2\exp (-\frac{32np_{n}\varepsilon ^{2}}{49}),\frac{14}{%
\varepsilon }\sqrt{\frac{V_{\mathcal{C}}(\ln (2(1-p_{n})+1)+4)}{n(1-p_{n})}}%
).
\end{equation*}%
to $\displaystyle2^{\frac{1}{p_{n}}}\exp \left( -\frac{n\epsilon ^{2}}{64(%
\sqrt{V\ln (2(2np_{n}+1))}+2)}\right) $. This result is important since it
shows why intensive use of the data can be very fruitful to improve the
estimation rate. Another interesting consequence of this proposition is
that, for a fixed precision $\varepsilon $, the size of the test is not
required to grow to infinity for the exponential convergence of the
cross-validation procedure. For this, it is sufficient that the size of the
test sample is larger than a fixed number $n_{0}$.

\bigskip \noindent {\bfseries  Proof.}
%\noindent As previously, we begin by splitting the probability according to $\widetilde {R}%
%_{n(1-p)}$, $\mathbb{P}(\widehat{R}_{CV}-\widetilde{R}_{n}\geq5 \varepsilon) \leq
%\mathbb{P}(\widehat{R}_{CV}-\widetilde{R}_{n(1-p)}\geq\varepsilon) + \mathbb{P}(\widetilde{R}_{n(1-p)}-\widetilde{R}_{n}\geq4\varepsilon)$.

%\noindent We have proved that $\mathbb{P}(\widetilde{R}_{n(1-p)}-\widetilde{R}_{n}\geq4\varepsilon)\leq 4(2n(1-p_n))^{\frac{4V}{1-p_n}} $

\bigskip

\noindent Recall that the size of the training sample is $n (1-p_{n})$, and
the size of the test sample is then $np_n$. For this proposition, we have $%
p_{n} < \frac{1}{2}$

\noindent We are interested in the behaviour of $\widehat{R}_{CV}-\bar{R}%
_{n(1-p)}=\mathbb{E}_{V_{n}^{tr}}\widehat{R}_{V_{n}^{ts}}(\phi
_{V_{n}^{tr}})-\mathbb{E}_{V_{n}^{tr}}R(\phi _{V_{n}^{tr}}$) which is a sum
of $\frac{1}{p_{n}}=k$ terms in the case of the $k$-fold cross-validation.

\noindent The difficulty is that these terms are neither
independent, nor even exchangeable. We have in mind to apply the
results about the sum of independent random variables. For this,
we need a way to introduce independence in our samples. In the
same time, we do not want to lose too much information. For this,
we will introduce independence by using by using the supremum. We
have,

\begin{equation*}
\displaystyle%
\begin{array}{rcl}
\Pr (\widehat{R}_{CV}-\bar{R}_{n(1-p)}\geq \varepsilon ) & = & \Pr (\mathbb{E%
}_{V_{n}^{tr}}(\widehat{R}_{V_{n}^{ts}}(\phi _{V_{n}^{tr}})-R(\phi
_{V_{n}^{tr}}))\geq \varepsilon ) \\
& \leq & \Pr (\mathbb{E}_{V_{n}^{tr}}(\sup_{\phi \in \mathcal{C}}(\widehat{R}%
_{V_{n}^{ts}}(\phi )-R(\phi ))\geq \epsilon ).%
\end{array}%
\end{equation*}

\noindent Now, we have a sum of $k=\frac{1}{p_{n}}$ i.i.d terms: $\mathbb{P}(%
\frac{1}{k}\sum Y_{i}\geq \epsilon)$, with $Y_{i}=\sup_{\phi\in\mathcal{C}}(%
\widehat{R}_{V_{n}^{ts}}(\phi)-R(\phi))$.

\noindent However, we have an extra piece of information: an upper bound for
the tail probability of these variables, using the concentration inequality
due to \cite{Vap98}.

\begin{equation*}
\Pr (\sup_{\phi \in \mathcal{C}}(\widehat{R}_{V_{n}^{ts}}(\phi )-R(\phi
))\geq \epsilon )\leq c(np_{n},V_{\mathcal{C}})e^{-\frac{\epsilon ^{2}}{%
2\sigma (np_{n})^{2}}}.
\end{equation*}

\noindent with $c(n,V_{\mathcal{C}})=2\mathcal{S}(2n,\mathcal{C}$ )$\leq
2(2n+1)^{V_{\mathcal{C}}}$ and $\sigma (n)^{2}=\frac{4}{n}$.

\noindent In fact, summing independent bounded variables with exponentially
small tail probability gives us a better concentration inequality than the
simple sum of independent bounded variables.

\noindent To show this, we proceed in three steps:

\begin{enumerate}
\item the $q$-H\"{o}lder norms of each variable is uniformly bounded by $%
\sqrt{q}$,

\item the Laplace transform of $Y_{i}$ is smaller than the Laplace transform
of some particular normal variable,

\item using Chernoff's method, we obtain a sharp concentration inequality.
\end{enumerate}

\begin{enumerate}
\item \noindent First step (for the proof, see appendices), we prove

\begin{lemma}
\label{lem:subgaussian}

\noindent Let $Y$ a random variable (bounded by $1)$ with subgaussian tail
probability $\mathbb{P(}Y\geq \varepsilon )\leq ce^{-\frac{\epsilon ^{2}}{%
2\sigma ^{2}}}$ for all $\varepsilon >0$ with $\sigma ^{2}>0$ and $c\geq 2$.
Then, there exists a constant $\gamma $ such that, for every integer $q$,
\begin{equation*}
(\mathbb{E}{Y_{+}}^{q})^{\frac{1}{q}}\leq \sqrt{\gamma q},
\end{equation*}%
\noindent with $\gamma =(\sigma \sqrt{4\ln (c)}+{\pi }^{\frac{1}{4}}3^{\frac{%
1}{3}}2e^{-\frac{1}{2}}\sigma )^{2}$.
\end{lemma}

\item Second step (see exercise 4 in \cite{Lug03}), we have

\begin{lemma}
\label{lem:holder}

If there exists a constant $\gamma $, such that for every integer $q$
\begin{equation*}
(\mathbb{E}{Y_{+}}^{q})^{\frac{1}{q}}\leq \sqrt{\gamma q}.
\end{equation*}%
\noindent then we have
\begin{equation*}
\mathbb{E}(e^{sY})\leq \sqrt{2}e^{\frac{1}{6}}e^{\frac{s^{2}e\gamma }{2}}.
\end{equation*}
\end{lemma}

\item Third step, we have the result using Chernoff's method.

\begin{lemma}
\label{lem:Chernoff}

If, for some $\alpha >0$, $\beta >0$, we have:
\begin{equation*}
\mathbb{E}(e^{sY})\leq \alpha e^{\frac{s^{2}\beta ^{2}}{2}}
\end{equation*}%
\noindent then if $(Y_{i})_{1\leq i\leq n}$ are i.i.d., we have:%
\begin{equation*}
\mathbb{P}(\frac{1}{V}\sum_{i=1}^{V}Y_{i}>\epsilon )\leq \alpha ^{V}e^{\frac{%
-V\epsilon ^{2}}{2\beta ^{2}}}
\end{equation*}
\end{lemma}
\end{enumerate}

\bigskip

\noindent Putting lemma \ref{lem:subgaussian} \ref{lem:holder} \ref%
{lem:Chernoff} together, we eventually get:

\begin{equation*}
\mathbb{P}(\mathbb{E}_{V_{n}^{tr}}(\sup_{\phi \in \mathcal{C}}\widehat{R}%
_{V_{n}^{tr}}(\phi )-R(\phi ))\geq \varepsilon )\leq (\sqrt{2}e^{1/6})^{%
\frac{1}{p_{n}}}\exp {\left( \frac{-\frac{1}{p_{n}}\epsilon ^{2}}{2\sigma
(np_{n})^{2}(e^{\frac{1}{2}}\sqrt{4\ln (c(np_{n},V_{\mathcal{C}}))}+{\pi }^{%
\frac{1}{4}}3^{\frac{1}{3}}2)^{2}}\right) .}
\end{equation*}

$\Box$

\noindent Symmetrically, we obtain:

\begin{proposition}[k-fold bis]
\noindent Suppose that $\mathcal{H}$ holds. Then, in the case of
the $k$-fold cross-validation procedure, we have for all
$\varepsilon>0$
$$
\mathbb{P}(\hat{R}_{n(1-p_n)} -\widehat{R}_{CV}
 \geq\varepsilon
)\leq 2^{\frac{1}{p_{n}}}\exp\left(-\frac{n\epsilon^{2}%
}{64(\sqrt{V_{\mathcal{C}}\ln(2(2np_n+1))}+2)}\right).%
$$
\end{proposition}

\noindent Eventually, we have a control on the absolute deviation

\begin{theorem}[Absolute error for the k-fold]
\label{ch1:thm kfold}
 Suppose that $\mathcal{H}$ holds. Then, in the case of
the $k$-fold cross-validation procedure, we have for all
$\varepsilon
>0$,
\begin{equation*}
\Pr (|\widetilde{R}_{n}-\widehat{R}_{CV}|\geq \varepsilon )\leq
B_{k}(n,p_{n},\varepsilon )+V_{k}(n,p_{n},\varepsilon )
\end{equation*}

with

\begin{itemize}
\item $B_{k}(n,p_{n},\varepsilon )=\displaystyle5(2n(1-1/k)+1)^{\frac{4V_{%
\mathcal{C}}}{1-1/k}}\exp (-\frac{n\varepsilon ^{2}}{64})$

\item { $V_{k}(n,p_{n},\varepsilon )=$
\begin{align*}
  & \min ( \exp (-%
\frac{2n/\varepsilon ^{2}}{25k}),\frac{16}{\varepsilon }\sqrt{\frac{V_{%
\mathcal{C}}(\ln (2(1-1/k)+1)+4)}{n(1-1/k)}}, \\
& \quad 22^{\frac{1}{p_{n}}}\exp(
-\frac{n\epsilon ^{2}}{25\ast 64(\sqrt{V_{\mathcal{C}}\ln (2(2np_{n}+1))}+2)}%
) ).
\end{align*}
}
\end{itemize}
\end{theorem}

\subsection{Hold-out cross-validation}

\noindent For hold-out cross-validation, the symmetric condition that for
all $i$, $\Pr \mathbb{(}i\in \mathcal{D}_{V_{n}^{tr}})$ is independent of $i$
is no longer valid. Indeed, in the hold-out cross-validation (or split
sample), there is no crossing again.

\bigskip \noindent In the next proposition, we suppose that the training
sample and the test sample are disjoint and that the number of observations
in the learning sample and in the test sample are still respectively $%
n(1-p_{n})$ and $np_{n}$. Moreover, we suppose also that the predictors $%
\phi _{n}$ are empirical risk minimizers on a class $\mathcal{C}$ with
finite $V_{\mathcal{C}}$-dimension $V_{\mathcal{C}}$ and $L$ a loss function
bounded by $1$. \textbf{We denote these hypotheses by $\mathcal{G}$.}

\bigskip \noindent We get the following result

\begin{theorem}[Hold-out]
\label{holdout}%
%\label{largets}%
\noindent Suppose that $\mathcal{G}$ holds. Then, we have for all $%
\varepsilon >0$,

{\small
\begin{equation*}
\Pr (|\widetilde{R}_{n}-\widehat{R}_{CV}|\geq \varepsilon )\leq
B_{hold}(n,p_{n},\varepsilon )+V_{hold}(n,p_{n},\varepsilon )
\end{equation*}
}

with

\begin{itemize}
\item $B_{hold}(n,p_{n},\varepsilon )=\displaystyle8(2n(1-p_{n})+1)^{4V_{%
\mathcal{C}}}\exp (-\frac{2n(1-p_{n})\varepsilon ^{2}}{25})$

\item $V_{hold}(n,p_{n},\varepsilon )=\displaystyle2\exp (-\frac{%
2np_{n}\varepsilon ^{2}}{25}).$
\end{itemize}
\end{theorem}

\noindent {\bfseries  Proof. } \noindent We just have to follow
the same steps as in proposition \ref{thm:sym}. But in the case
of hold-out cross-validation, notice that

\begin{align*}
\Pr \mathbb{(E}_{V_{n}^{tr}}\sup_{\phi \in \mathcal{C}}(\widehat{R}%
_{V_{n}^{tr}}(\phi )-R(\phi ))\geq \varepsilon )& =\Pr (\sup_{\phi \in
\mathcal{C}}(\widehat{R}_{\mathbf{v}_{n}^{tr}}(\phi )-R(\phi ))\geq
\varepsilon ) \\
& \leq \mathcal{S}(2n(1-p_{n}),\mathcal{C})^{4}e^{-n(1-p_{n})\varepsilon
^{2}}
\end{align*}

Moreover, the lemma \ref{ch1:monlemme} is no longer valid, since $\mathbb{E}%
_{V_{n}^{tr}}R_{V_{n}^{ts}}(\phi _{n})\neq \widehat{R}_{n}$. \bigskip $\Box $

\subsection{Discussion}

We base the next discussion on upperbounds, so the following
heuristic arguments are questionable if the bounds are loose.

\subsubsection*{Crossing versus non-crossing}

\noindent One can wonder: what is the use of averaging again over the
different folds of the $k$-fold cross-validation, which is time consuming?
As far as the expected errors are concerned, the upper bounds are the same
for crossing cross-validation procedures and for hold-out cross-validation.
But suppose we are given a level of precision $\varepsilon $, and we want to
find an interval of length $2\varepsilon $ with maximal confidence. Then
notice that $B_{sym}/B_{hold}=(2n(1-p_{n})+1)^{\frac{4V_{\mathcal{C}}p_{n}}{%
1-p_{n}}}\exp (-np_{n}\varepsilon ^{2})$. Thus if $p_{n}$ is constant, $%
B_{sym}/B_{hold}\rightarrow _{n\rightarrow \infty }0$: the term $B$ will be
much greater for hold-out based on large learning size. On the contrary, if
the learning size is small, then the term $B$ is smaller for non crossing
procedure for a given $p_{n}$. This might due to the absence of resampling.

\noindent Regarding the variance term $V_{hold}(n,p_{n},\varepsilon )$, we
need the size of the test sample to grow to infinity for the consistency of
the hold-out cross-validation. On the contrary, for crossing
cross-validation, the term $V$ converges to $0$ whatever the size of the
test is.

\subsubsection*{$k$-fold cross-validation versus others}

\noindent If we consider the $L_{1}$ error, the upper bounds are
the same for crossing cross-validation procedures and for other
cross-validation procedures. But if we look for the interval of
length $2\varepsilon $ with maximal confidence, then notice that
$V_{k}/V_{sym}\rightarrow _{n\rightarrow \infty }0$ (with
$V_k,V_{sym}$ defined respectively in theorems \ref{ch1:thm
kfold}, \ref{thm:sym})
if the number of elements in the training sample $%
np_{n}$ is constant and large enough. Thus, if the learning size is large
enough, the $V$ term is much smaller for the $k$-fold cross-validation,
thanks to the crossing.

\subsubsection*{Estimation curve}

\noindent The expression of the variance term $V$ depends on the percentage
of observations $p_{n}$ in the test sample and on the type of
cross-validation procedure. We have thus a control of the variance term
depending on $p_{n}.$

\begin{figure}[tbh]
\begin{center}
\includegraphics[angle=0,scale=0.6]{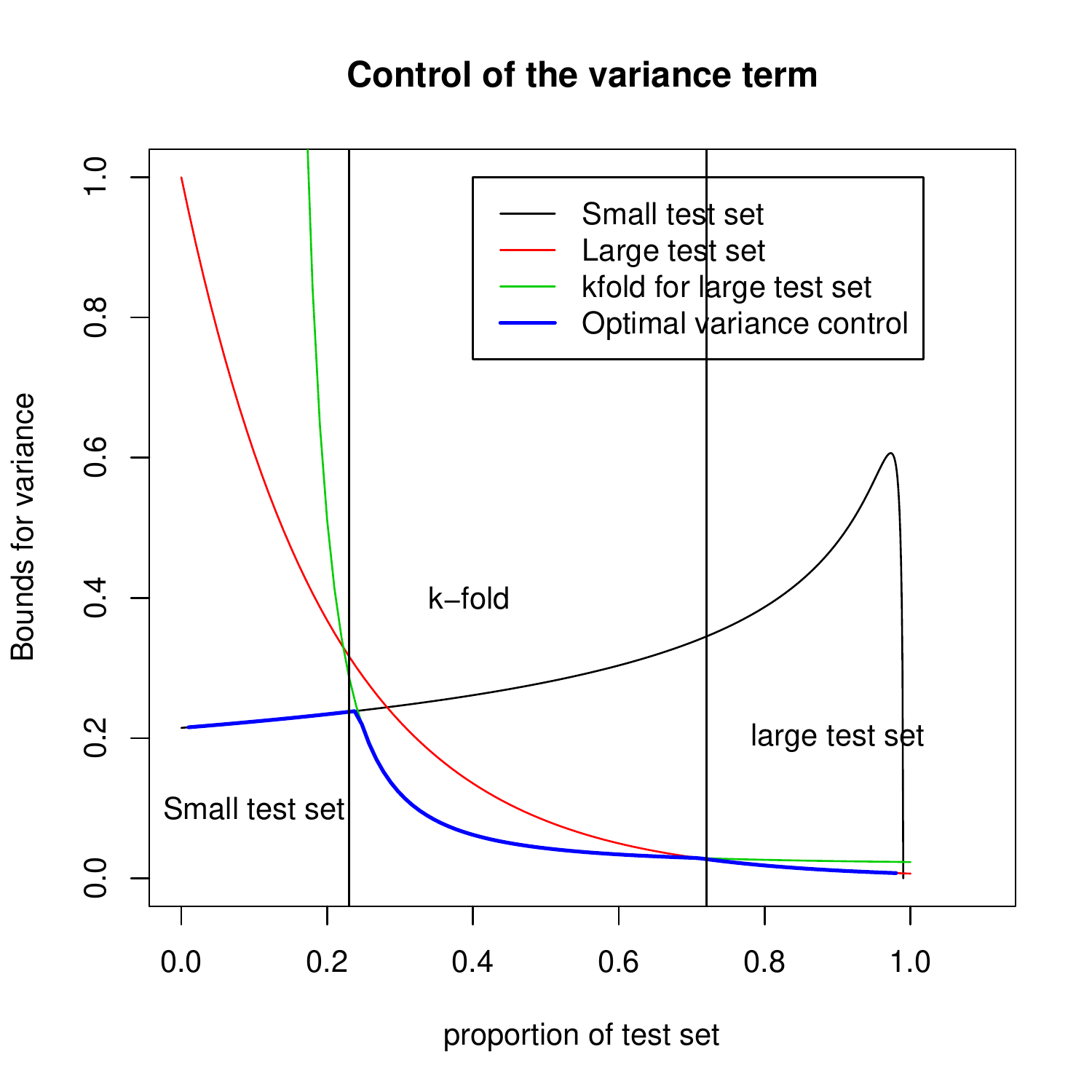}
\end{center}
\par
%\caption{Terminology for the datasamples used in cross-validation.}%
\end{figure}

\noindent We can define the estimation curve (in probability or in $L_{1}$
norm) which gives for each cross-validation procedure and for each $p_{n}$
the estimation error.

\begin{definition}[Estimation curve in probability]
Let $\varepsilon >0$:
\begin{equation*}
\mathcal{AC}:p_{n}\mapsto B(n,p_{n},\varepsilon )+V(n,p_{n},\varepsilon ).
\end{equation*}%
with $B(n,p_{n},\varepsilon )$ and $V(n,p_{n},\varepsilon )$ defined in
theorem \ref{thm:sym}.
\end{definition}

\noindent This can be done with the expectation of the absolute of deviation
or with the probability upper bound if the level of precision is $%
\varepsilon $.

\begin{definition}[Estimation curve in $L_{1}$ norm]
\begin{equation*}
\mathcal{AC}:p_{n}\mapsto B(n,p_{n})+V(n,p_{n}).
\end{equation*}%
with $B(n,p_{n})$ and $V(n,p_{n})$ defined as in proposition \ref%
{cor:l1-largets}.
\end{definition}

\noindent We say that the estimation curve in probability
experiences a  phase transition when the convergence rate
$V(n,p_{n},\varepsilon)$\ changes. The estimation curve
experiences at least one transition phase. The transition phases
just depend on the class of predictors and on the sample size. On
the contrary of the learning curve, the transition phases of the
estimation curve are independent of the underlying distribution.
The
different transition phases define three different regions in the values of $%
p_{n}$ the percentage of observations in the test sample. This three regions
emphasize the different roles played by small test sample cross-validation,
large test samples cross-validation and $k$-fold cross-validation.

\subsubsection*{Optimal splitting and confidence intervals}

\noindent The estimation curve gives a hint for this simple but important
question: how should one choose the cross-validation procedure in order to
get the best estimation rate? How should one choose $k$ in the $k$-fold
cross-validation? The quantitative answer of theses questions is the $%
\arg\min$ of the estimation curve $\mathcal{AC}$.

\noindent That is in probability

\begin{equation*}
p_{n}^{\star }(\varepsilon )=\arg \min_{p_{n}}\mathcal{AC}(p_{n},\varepsilon
).
\end{equation*}

\noindent or in $L_1$ norm:

\begin{equation*}
p_{n}^{\star }=\arg \min_{p_{n}}\mathcal{AC}(p_{n}).
\end{equation*}

\noindent As far as the $L_{1}$ norm is concerned, we can derive
a simple expression for the choice of $p_{n}$. Indeed, if we use
chaining arguments in the proof of proposition \ref{ch1:lemme1},
that is: there exists a universal
constant $c>0$ such that $\mathbb{E}\sup_{\phi \in \mathcal{C}}(\widehat{R}_{%
\mathbf{W}_{n}^{tr}}(\phi )-R(\phi ))\leq c\sqrt{\frac{V_{\mathcal{C}}}{%
n(1-p_{n})}}$ (for the proof, see e.g. \cite{DGL96}). The proposition \ref%
{cor:l1-largets} thus becomes:

\begin{corollary}[$L_{1}$ error for large test sample]
Suppose that $\mathcal{H}$ holds. Then, there exists a universal constant $%
c>0$ such that:
\begin{equation*}
\mathbb{E}|\widehat{R}_{CV}-\widetilde{R}_{n}|\leq c\sqrt{\frac{V_{\mathcal{C%
}}}{n(1-p_{n})}}+2\sqrt{\frac{6}{np_{n}}}.
\end{equation*}
\end{corollary}

\noindent We can then minimize the last expression in $p_{n}$. After
derivation, we obtain $p_{n}^{\star }=((\frac{c^{2}V_{\mathcal{C}}}{2\sqrt{(}%
6)})^{1/3}+1)^{-1}$. Thus, the larger the VC-dimension is, the larger the
training sample should be. Since it may be difficult to find an explicit
constant, one may try to solve: $\sqrt{\frac{V_{\mathcal{C}}(\ln (2n)+4))}{%
n(1-p_{n})}}+2\sqrt{\frac{6}{np_{n}}}$. We obtain then a computable rule $%
p_{n}^{\star }=((\frac{V_{\mathcal{C}}(\ln (2n)+4))}{2\sqrt{(}6)}%
)^{1/3}+1)^{-1}$

\bigskip

\begin{figure}[tbh]
\begin{center}
\includegraphics[angle=0,scale=0.6]{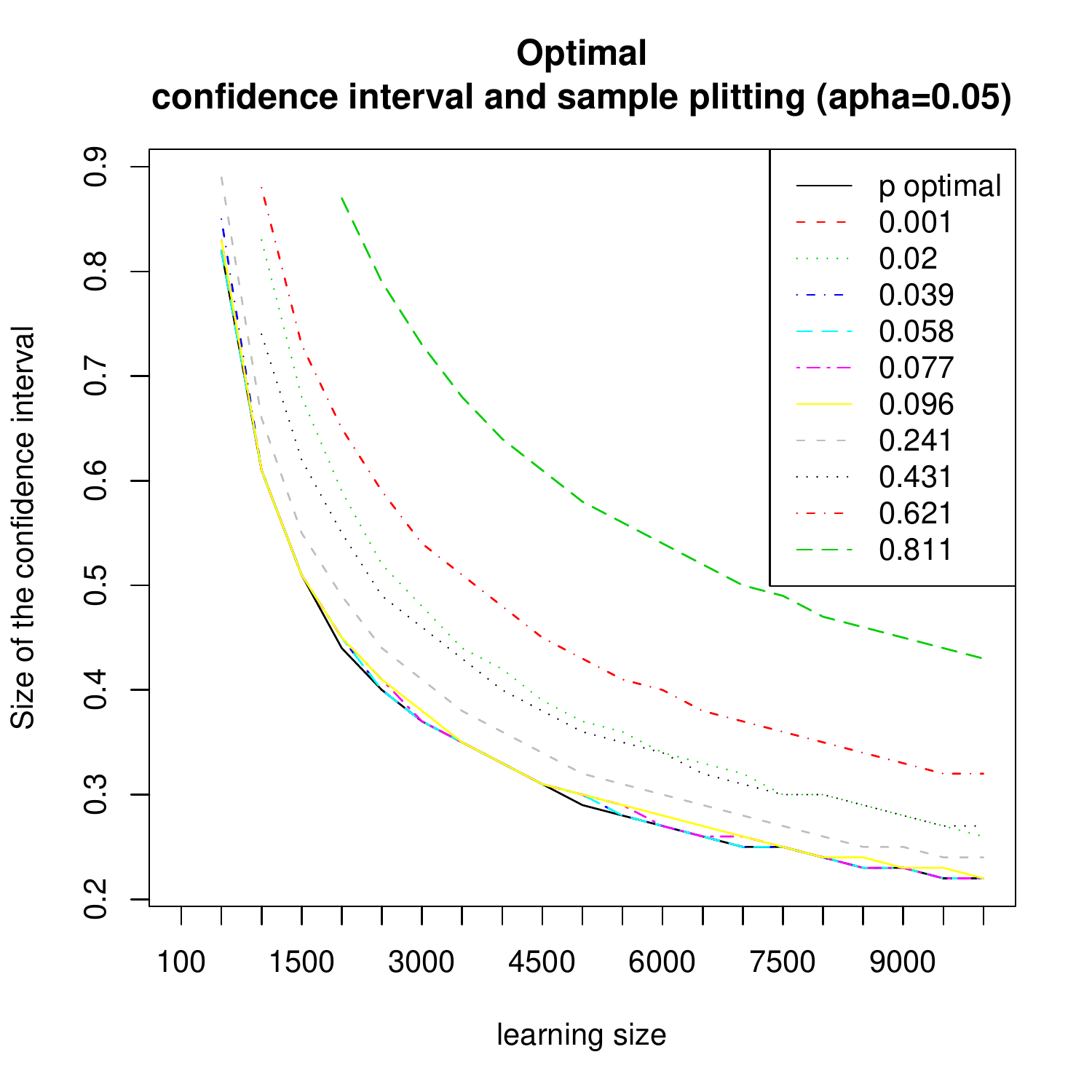}
\end{center}
\caption{Upperbounds for cross-validation procedures with different
splitting }
\label{fig:Splitting}
\end{figure}

\noindent Another interesting issue is: knowing the number of observations $%
n $ and the class of predictors, we can now derive an optimal minimal $%
1-\alpha $-confidence interval, together with the cross-validation
procedure. We look at the values $(\varepsilon ,p_{n})$ such that the
upperbound $B(n,p_{n},\varepsilon )+V(n,p_{n},\varepsilon )$ is below the
threshold $\alpha $. Then, we select the couple $(\varepsilon ^{\ast
},p_{n}^{\ast })$ among those values for which $\varepsilon $ is minimal. On
figure \ref{fig:Splitting}, we fix a choice of $\alpha =5\%$. We observe
that, for values of $n$ between $1000$ and $10000$ and for small
VC-dimension, a choice of $p\simeq 10\%$, i.e. the ten-fold
cross-validation, seems to be a reasonable choice.

\newpage

\section{Appendices}

\subsection{Notations and definitions}

\noindent We recall the main notations and definitions.

\begin{table}[th]
\begin{center}
\begin{tabular}{lll}
\hline
Name & Notation & Definition \\ \hline\hline
Generalisation error & $\widetilde{R}_{n}$ & $\mathbb{E}_{P}[L(Y,\phi (X,%
\mathcal{D}_{n}))\mid \mathcal{D}_{n}]$ \\
Resubstitution estimate & $\widehat{R}_{n}$ & $\frac{1}{n}%
\sum_{i=1}^{n}L(Y_{i},\phi _{n}(X_{i},\mathcal{D}_{n}))$ \\
Cross-validation estimate & $\widehat{R}_{CV}$ & $\mathbb{E}_{V_{n}^{tr}}%
\hat{R}_{V_{n}^{ts}}(\phi _{V_{n}^{tr}})$ \\
Cross-validation risk & $\bar{R}_{n(1-p)}$ & $\mathbb{E}_{V_{n}^{tr}}R(\phi
_{V_{n}^{tr}})$ \\
Optimal error & $R_{opt}$ & $\inf_{\phi \in \mathcal{C}}R(\phi )$ \\ \hline
\end{tabular}%
\end{center}
\caption{Main notations}
\end{table}

\subsection{Proofs}

\noindent We recall three very useful results. The first one, due to %
\cite{HOEF63}, bounds the difference between the empirical mean
and the expected value. The second one, due to \cite{VC71} ,
bounds the supremum over the class of predictors of the
difference between the training error and the generalization
error. The last one is called the bounded differences inequality
\cite{McD89} .

\begin{theorem}[\cite{HOEF63}]
Let $X_{1,\ldots,}X_{n}%
$ independent random
variables in $[a_{i},b_{i}%
]$.
Then for all $\varepsilon>0,$%
\[
\mathbb{P}(\sum X_{i}-\mathbb{E}%
(\sum X_{i})\geq n\mathbb{\varepsilon})\leq
e^{-\frac{2\varepsilon^{2}}%
{\sum_{i}(b_{i-}a_{i})^{2}}}%
\]
\end{theorem}

\begin{theorem}
[\cite{VC71}] Let $\mathcal{C}%
$ a class of predictors with finite VC-dimension and $L$ a loss function bounded by $1$. Then for all $\varepsilon>0,$%
$$ \mathbb{P}%
(\sup_{\phi\in\mathcal{C}}(\widehat{R}_{n}%
(\phi)-L(\phi
))\geq\varepsilon)\leq c(n,V_{\mathcal{C}})e^{-\frac{\varepsilon^{2}%
}{2\sigma(n)^{2}}}$$
with $c(n,V_{\mathcal{C}})=2\mathcal{S}(2n,\mathcal{C}%
$)$\leq2(2n+1)^{V_{\mathcal{C}}}$ and if $n\geq
V_{\mathcal{C}},$ $2\mathcal{S}(2n,\mathcal{C}%
$)$\leq2(\frac{2ne}{V_{\mathcal{C}}})^{V_{\mathcal{C}}}$ and
$\sigma(n)^{2}=\frac{4}{n}$
\end{theorem}

\begin{theorem}[McDiarmid]
Let $X_{1,\ldots,}X_{n}%
$ be independent random variables taking values in a
sample $\mathcal{X}$, and assume that $f:\mathcal{X}^{n}%
\rightarrow \mathcal{R}$ satisfies%
\[
\forall i,\sup_{\substack{x_{1}%
,\ldots,x_{i},\ldots,x_{n} \\x_{i}^{^{\prime}}%
}}|f(x_{1},...,x_{n}%
)-f(x_{1},...,x_{i^{\prime}},...,x_{n})|\leq c_{i}.%
\]
Then, for all $\varepsilon>0,$%
\[
\mathbb{P(}f(X_{1},...,X_{n}%
)-\mathbb{E}f(X_{1},...,X_{n})\geq\varepsilon)\leq
e^{-\frac{2\varepsilon^{2}%
}{\sum_{i}c_{i}^{2}}}.%
\]
\end{theorem}

\bigskip

\subsubsection{\bfseries  Proof of lemma \ref{ch1:lemme1}}

\noindent First, notice that

\begin{equation*}
\mathbb{P(E}_{V_{n}^{tr}}\sup_{\phi \in \mathcal{C}}(\widehat{R}%
_{V_{n}^{tr}}(\phi )-R(\phi ))-\mathbb{EE}_{V_{n}^{tr}}\sup_{\phi \in
\mathcal{C}}(\widehat{R}_{V_{n}^{tr}}(\phi )-R(\phi ))\geq \varepsilon )\leq
e^{-2n\varepsilon ^{2}},
\end{equation*}

\noindent using McDiarmid's inequality by setting $f(X_{1},\ldots ,X_{n})=%
\mathbb{E}_{V_{n}^{tr}}\sup_{\phi \in \mathcal{C}}(\widehat{R}%
_{V_{n}^{tr}}(\phi )-R(\phi ))$ and since for all $i$,

\begin{align*}
& \sup_{\substack{ x_{1},\ldots ,x_{i},\ldots ,x_{n}  \\ x_{i}^{^{\prime }}}}%
|\mathbb{E}_{V_{n}^{tr}}\sup_{\phi \in \mathcal{C}}(\widehat{R}%
_{V_{n}^{tr}}(\phi )-R(\phi ))-\mathbb{E}_{V_{n}^{tr}}\sup_{\phi \in
\mathcal{C}}(\widehat{R}_{V_{n}^{tr}}^{^{\prime }}(\phi )-R(\phi ))| \\
& =\sup_{\substack{ x_{1},\ldots ,x_{i},\ldots ,x_{n}  \\ x_{i}^{^{\prime }}
}}\left| \mathbb{E}_{V_{n}^{tr}}\left[ \sup_{\phi \in \mathcal{C}}(\widehat{R%
}_{V_{n}^{tr}}(\phi )-R(\phi ))-\sup_{\phi \in \mathcal{C}}(\widehat{R}%
_{V_{n}^{tr}}^{^{\prime }}(\phi )-R(\phi ))\right] \right| \\
& \leq \sup_{\substack{ x_{1},\ldots ,x_{i},\ldots ,x_{n}  \\ %
x_{i}^{^{\prime }}}}\mathbb{E}_{V_{n}^{tr}}\left| \sup_{\phi \in \mathcal{C}%
}(\widehat{R}_{V_{n}^{tr}}(\phi )-R(\phi ))-\sup_{\phi \in \mathcal{C}}(%
\widehat{R}_{V_{n}^{tr}}^{^{\prime }}(\phi )-R(\phi ))\right| \\
& \text{by Jensen's inequality} \\
& \leq \sup_{\substack{ x_{1},\ldots ,x_{i},\ldots ,x_{n}  \\ %
x_{i}^{^{\prime }}}}\mathbb{E}_{V_{n}^{tr}}\sup_{\phi \in \mathcal{C}}|%
\widehat{R}_{V_{n}^{tr}}(\phi )-\widehat{R}_{V_{n}^{tr}}^{^{\prime }}(\phi )|
\\
& \text{since }|\sup f-\sup g|\leq \sup |f-g| \\
& \leq \frac{1}{n}.
\end{align*}

\noindent Indeed, if we note $Q$ the number of elements in the sum $\mathbb{E%
}_{V_{n}^{tr}}$, the number of changes is lower than $\leq \frac{1}{Q}(\frac{%
1}{n(1-p_{n})}\text{multiplied by the number of times }i^{\prime
}\text{ in
the learning sample})$ that is $\frac{1}{Q}(\frac{1}{n(1-p_{n})}Q(1-p_{n}))=%
\frac{1}{n}$

 \noindent Furthermore, we have

\begin{align*}
\mathbb{EE}_{V_{n}^{tr}}\sup_{\phi \in \mathcal{C}}(\widehat{R}%
_{V_{n}^{tr}}(\phi )-R(\phi ))& =\mathbb{E}\sup_{\phi \in \mathcal{C}}(%
\widehat{R}_{\mathbf{v}_{n}^{tr}}(\phi )-R(\phi )) \\
& \text{with }\mathbf{v}_{n}^{tr}\text{ a fixed vector} \\
& \leq \sqrt{\frac{2\ln (\mathcal{S}(2n(1-p_{n}),\mathcal{C})}{n(1-p_{n})}.}
\end{align*}

\noindent by Vapnik-Chernovenkis's inequality.

\bigskip

\noindent Thus, if we denote $\Pr
\mathbb{(E}_{V_{n}^{tr}}\sup_{\phi \in
\mathcal{C}}(\widehat{R}_{V_{n}^{tr}}(\phi )-R(\phi ))\geq
\varepsilon )$ by $P_{1}$ it leads to
\begin{align*}
P_{1}=& \Pr \mathbb{(E}_{V_{n}^{tr}}\sup_{\phi \in \mathcal{C}}(\widehat{R}%
_{V_{n}^{tr}}(\phi )-R(\phi ))-\mathbb{EE}_{V_{n}^{tr}}\sup_{\phi \in
\mathcal{C}}(\widehat{R}_{V_{n}^{tr}}(\phi )-R(\phi )) \\
& \qquad \geq \varepsilon -\mathbb{EE}_{V_{n}^{tr}}\sup_{\phi \in \mathcal{C}%
}(\widehat{R}_{V_{n}^{tr}}(\phi )-R(\phi )).
\end{align*}

\noindent Then, using the two previous inequalities
\begin{align*}
P_{1}\leq & \Pr \mathbb{(E}_{V_{n}^{tr}}\sup_{\phi \in \mathcal{C}}(\widehat{%
R}_{V_{n}^{tr}}(\phi )-R(\phi ))-\mathbb{EE}_{V_{n}^{tr}}\sup_{\phi \in
\mathcal{C}}(\widehat{R}_{V_{n}^{tr}}(\phi )-R(\phi )) \\
& \qquad \geq \varepsilon -\sqrt{\frac{2\ln (\mathcal{S}(2n(1-p_{n}),%
\mathcal{C})}{n(1-p_{n})}}).
\end{align*}

\noindent Since $2(u-v)^{2}\geq u^{2}-2v^{2}$, it follows
\begin{align*}
P_{1}\leq & \exp (-2n(\varepsilon -\sqrt{\frac{2\ln (\mathcal{S}(2n(1-p_{n}),%
\mathcal{C})}{n(1-p_{n})}})^{2})\leq \exp (-n(\varepsilon ^{2}-\frac{4\ln (%
\mathcal{S}(2n(1-p_{n}),\mathcal{C})}{n(1-p_{n})}))) \\
\leq & \mathcal{S}(2n(1-p_{n}),\mathcal{C})^{4/(1-p_{n})}\exp (-n\varepsilon
^{2}).
\end{align*}

$\Box$ %Control the holder norms for subgaussian distributions

%\bigskip
\bigskip

\subsubsection{{\bfseries  Proof of lemma \ref{ch1:lemme1}}}

\noindent Recall that $\widehat{R}_{CV}=\mathbb{E}_{V_{n}^{tr}}\widehat{R}%
_{_{V_{n}^{ts}}}(\phi _{V_{n}^{tr}})$

\bigskip

\noindent But by definition of $\phi_{n}$, we have $\widehat{R}%
_{n}(\phi_{n}) \leq\widehat{R}_{n}(\phi_{V_{n}^{tr}})$.

\bigskip

\noindent It follows that $\frac{1}{n}(np_{n}\widehat{R}_{_{V_{n}^{ts}}}(%
\phi _{n})+\sum_{i\in V_{n}^{tr}}L(Y_{i,}\phi _{n}(X_{i}))\leq \frac{1}{n}%
(np_{n}\widehat{R}_{_{V_{n}^{ts}}}(\phi _{V_{n}^{tr}})\quad +\sum_{i\in
V_{n}^{tr}}L(Y_{i,}\phi _{V_{n}^{tr}}(X_{i})).$

\bigskip \noindent Thus, since $\sum_{i\in V_{n}^{tr}}L(Y_{i,}\phi
_{n}(X_{i}))\geq \sum_{i\in V_{n}^{tr}}L(Y_{i,}\phi _{V_{n}^{tr}}(X_{i}))$
by definition of $\phi _{V_{n}^{tr}}$, we have $\widehat{R}%
_{_{V_{n}^{ts}}}(\phi _{n})\leq \widehat{R}_{_{V_{n}^{ts}}}(\phi
_{V_{n}^{tr}}).$ \bigskip

\noindent From this, we deduce $\widehat{R}_{CV}=\mathbb{E}_{V_{n}^{tr}}%
\widehat{R}_{_{V_{n}^{ts}}}(\phi _{V_{n}^{tr}})\geq \mathbb{E}_{V_{n}^{tr}}%
\widehat{R}_{_{V_{n}^{ts}}}(\phi _{n})=\widehat{R}_{n}.$

$\Box$ \bigskip

\subsubsection{{\bfseries  Proof. of lemma \ref{ch1:monlemme}}}

\begin{equation*}
\forall \varepsilon >0,\mathbb{P}(X\geq \varepsilon )\leq \mathbb{P}%
(X_{+}\geq \varepsilon )\leq \frac{\mathbb{E}X_{+}}{\varepsilon }=\frac{%
\mathbb{E}X_{\_}}{\varepsilon }=\frac{\int_{0}^{1}\mathbb{P}(X_{-}\geq x)dx}{%
\varepsilon }=\frac{\int_{0}^{1}\mathbb{P}(X\leq -x)dx}{\varepsilon }.
\end{equation*}

$\Box$ \bigskip

\subsubsection{{\bfseries  Proof. of lemma \ref{lem:subgaussian}}}

\noindent First, suppose that $q>1$ and notice that
\begin{equation*}
\displaystyle%
\begin{array}{rcl}
\mathbb{E}{Y_{+}}^{q} & = & \int_{0}^{\infty }qy^{q-1}\mathbb{P}(Y_{+}>y)dy
\\
& = & q\int_{0}^{\infty }y^{q-1}\mathbb{P}(Y>y)dy.%
\end{array}%
\end{equation*}

\noindent We thus deduce that because of the subgaussian inequality:

\begin{equation*}
\displaystyle%
\begin{array}{rcl}
\mathbb{E}{Y_{+}}^{q} & \leq & q\int_{0}^{\sigma \sqrt{4\ln (c)}%
}y^{q-1}dy+q\int_{\sigma \sqrt{4\ln (c)}}^{\infty }cy^{q-1}e^{-\frac{y^{2}}{%
2\sigma ^{2}}}dy.%
\end{array}%
\end{equation*}

\noindent Then, with $\mathcal{N}$ a standard normal:
\begin{equation*}
\displaystyle%
\begin{array}{rcl}
\mathbb{E}{Y_{+}}^{q} & \leq & (\sigma \sqrt{4\ln (c)})^{q}+qc\int_{\sigma
\sqrt{4\ln (c)}}^{\infty }y^{q-1}e^{-\frac{y^{2}}{2\sigma ^{2}}}dy \\
& \leq & (\sigma \sqrt{4\ln (c)})^{q}+qc\sqrt{2\pi }\sigma \mathbb{E}%
((\sigma \mathcal{N})^{q-1}1_{(\sigma \sqrt{4\ln (c)}\leq \sigma \mathcal{N}%
)}).%
\end{array}%
\end{equation*}

\noindent This gives by Cauchy-Schwarz's inequality:

\begin{equation*}
\displaystyle%
\begin{array}{rcl}
\mathbb{E}{Y_{+}}^{q} & \leq & (\sigma \sqrt{4\ln (c)})^{q}+qc\sqrt{2\pi }%
\sigma ^{q}(\mathbb{E}\mathcal{N}^{2(q-1)}1_{0\leq \mathcal{N}})^{\frac{1}{2}%
}(\mathbb{P}(\sqrt{4\ln (c)}\leq \mathcal{N}))^{\frac{1}{2}}.%
\end{array}%
\end{equation*}

\noindent It leads to, since $\mathbb{E}\mathcal{N}^{2p}=\frac{(2p)!}{2^{p}p!%
}$, and$\sqrt{4\ln (c)}\geq 1$,
\begin{equation*}
\displaystyle%
\begin{array}{rcl}
\mathbb{E}{Y_{+}}^{q} & \leq & (\sigma \sqrt{4\ln (c)})^{q}+(2\pi
)^{1/4}qc\sigma ^{q}(\mathbb{E}\mathcal{N}^{2(q-1)})^{\frac{1}{2}}(e^{-\frac{%
(2)\ln (c)}{2}})^{\frac{1}{2}} \\
& \leq & (\sigma \sqrt{4\ln (c)})^{q}+(2\pi )^{1/4}q\sigma ^{q}(\frac{%
(2(q-1))!}{2^{(q-1)}(q-1)!})^{\frac{1}{2}}.%
\end{array}%
\end{equation*}

\noindent We obtain, since $\sqrt{2\pi n}(\frac{n}{e})^{n}e^{\frac{1}{12n}%
}\leq n!\leq \sqrt{2\pi n}(\frac{n}{e})^{n}e^{\frac{1}{12n+1}}$,

\begin{equation*}
\displaystyle%
\begin{array}{rcl}
\mathbb{E}{Y_{+}}^{q} & \leq & (\sigma \sqrt{4\ln (c)})^{q}+(2\pi
)^{1/4}q\sigma ^{q}\left( \frac{\sqrt{2\pi 2(q-1)}(\frac{2(q-1)}{e}%
)^{2(q-1)}e^{\frac{1}{24(q-1)+1}}}{2^{(q-1)}\sqrt{2\pi (q-1)}(\frac{(q-1)}{e}%
)^{k(q-1)}e^{\frac{1}{12(q-1)}}}\right) ^{\frac{1}{2}} \\
& \leq & (\sigma \sqrt{4\ln (c)})^{q}+(2\pi )^{1/4}q\sigma ^{q}(\sqrt{2}(%
\frac{2(q-1)}{e})^{(q-1)}e^{\frac{1}{24(q-1)+1}-\frac{1}{12(q-1)}})^{\frac{1%
}{2}} \\
& \leq & (\sigma \sqrt{4\ln (c)})^{q}+(2\pi )^{1/4}q2^{\frac{1}{4}}\sigma
^{q}(\frac{2(q-1)}{e})^{\frac{q-1}{2}}.%
\end{array}%
\end{equation*}

\noindent Thus, since $(a+b)^{\frac{1}{q}} \leq a^{\frac{1}{q}} + b^{\frac{1%
}{q}}, a,b \geq0$:

\begin{equation*}
\displaystyle%
\begin{array}{rcl}
(\mathbb{E}{Y_{+}}^{q})^{\frac{1}{q}} & \leq & \left( (\sigma \sqrt{4\ln (c)}%
)^{q}+(2\pi )^{1/4}q2^{\frac{1}{4}}\sigma ^{q}(\frac{2(q-1)}{e})^{\frac{q-1}{%
2}}\right) ^{\frac{1}{q}} \\
& \leq & \left( (\sigma \sqrt{4\ln (c)})^{q}\right) ^{\frac{1}{q}}+\left(
(2\pi )^{1/4}q2^{\frac{1}{4}}\sigma ^{q}(\frac{2(q-1)}{e})^{\frac{q-1}{2}%
}\right) ^{\frac{1}{q}},%
\end{array}%
\end{equation*}

\noindent which gives since $q^{\frac{1}{q}}\leq 3^{\frac{1}{3}},(\frac{%
2(q-1)}{e})^{\frac{q-1}{2q}}\leq (\frac{2q}{e})^{\frac{q-1}{2q}}\leq (\frac{%
2q}{e})^{\frac{q}{2q}}$ since $\frac{2q}{e}\geq 1$:
\begin{equation*}
\displaystyle%
\begin{array}{rcl}
(\mathbb{E}{Y_{+}}^{q})^{\frac{1}{q}} & \leq & \sigma \sqrt{4\ln (c)}+q^{%
\frac{1}{q}}((2\pi )^{1/4}2^{\frac{1}{4}})^{\frac{1}{q}}\sigma (\frac{2(q-1)%
}{e})^{\frac{q-1}{2q}} \\
& \leq & \sigma \sqrt{4\ln (c)}+3^{\frac{1}{3}}2^{\frac{1}{4}}\sigma (\frac{%
2q}{e})^{\frac{1}{2}}.%
\end{array}%
\end{equation*}

\begin{equation*}
\displaystyle%
\begin{array}{rcl}
& \leq & \sigma \sqrt{4\ln (c)}+(2\pi )^{1/4}3^{\frac{1}{3}}2^{\frac{3}{4}%
}e^{-\frac{1}{2}}\sigma \sqrt{q} \\
& \leq & (\sigma \sqrt{4\ln (c)}+(2\pi )^{1/4}3^{\frac{1}{3}}2^{\frac{3}{4}%
}e^{-\frac{1}{2}}\sigma )\sqrt{q} \\
& \leq & \sqrt{\gamma q}.%
\end{array}%
\end{equation*}

\noindent with $\gamma=(\sigma\sqrt{4\ln(c)}+(2 \pi)^{1/4} 3^{\frac{1}{3}}2^{%
\frac{3}{4}}e^{-\frac{1}{2}}\sigma)^{2}$

\noindent For $q=1$, notice that:
\begin{equation*}
\displaystyle%
\begin{array}{rcl}
(\mathbb{E}{Y_{+}}^{q})^{\frac{1}{q}} & \leq & \sigma \sqrt{4\ln (c)}+\frac{1%
}{2}\sigma \\
& \leq & \sqrt{\gamma q}.%
\end{array}%
\end{equation*}

$\Box$

\end{document}